\documentclass[twoside]{article}

\usepackage[accepted]{aistats2021}

\usepackage[round]{natbib}

\usepackage{mathtools}
\usepackage{bbm}
\usepackage{amsfonts}
\usepackage{subcaption}
\usepackage{textcomp}
\usepackage{multirow}
\usepackage[hidelinks]{hyperref}
\usepackage{tikz}
\usetikzlibrary{bayesnet}
\usetikzlibrary{arrows}
\usetikzlibrary{arrows.meta}
\usetikzlibrary{calc}

\newcommand{\kl}[2]{\operatorname{KL}({#1} \; || \; {#2})}
\newcommand{\E}[2]{\mathbb{E}_{#1} \left[#2\right]}
\newcommand{\ENB}[1]{\mathbb{E}_{#1}}  
\newcommand{\Norm}{\mathcal{N}}

\DeclareMathOperator*{\argmax}{arg\,max}

\newcommand{\modelname}{PartedVAE}

\begin{document}

\twocolumn[
\aistatstitle{Semi-Supervised Disentanglement of Class-Related and Class-Independent Factors in VAE}

\runningauthor{Sina Hajimiri, Aryo Lotfi, Mahdieh Soleymani Baghshah}
\aistatsauthor{ 
    Sina Hajimiri \\ Sharif University of Technology \\ \texttt{sihajimiri@ce.sharif.edu} \And 
    Aryo Lotfi \\ Sharif University of Technology \\ \texttt{arlotfi@ce.sharif.edu} \And 
    Mahdieh Soleymani Baghshah \\ Sharif University of Technology \\ \texttt{soleymani@sharif.edu} }
\aistatsaddress{}
]

\begin{abstract}
    In recent years, extending variational autoencoder's framework to learn disentangled representations has received much attention. We address this problem by proposing a framework capable of disentangling class-related and class-independent factors of variation in data. Our framework employs an attention mechanism in its latent space in order to improve the process of extracting class-related factors from data. We also deal with the multimodality of data distribution by utilizing mixture models as learnable prior distributions, as well as incorporating the Bhattacharyya coefficient in the objective function to prevent highly overlapping mixtures. Our model's encoder is further trained in a semi-supervised manner, with a small fraction of labeled data, to improve representations' interpretability. Experiments show that our framework disentangles class-related and class-independent factors of variation and learns interpretable features. Moreover, we demonstrate our model's performance with quantitative and qualitative results on various datasets.
\end{abstract}

\section{INTRODUCTION}

Representation learning is an important problem in machine learning that has an essential role in analyzing high dimensional data. In the past few years, there has been a surge of interest in learning disentangled representation, which attempts to discover distinct factors of variation in data.
In a disentangled representation, altering a single unit of the representation changes only one of the data's variation factors and does not affect other factors of variation (\cite{Bengio2013RepresentationLA, jointvae}). 
Variational autoencoder (VAE) (\cite{vae, Rezende2014StochasticBA}), a deep generative model capable of learning representations of data, has been a popular framework for learning disentangled representations (\cite{Tschannen2018RecentAI, betavae, factorvae, betatcvae, jointvae, guided}).

In most real-world datasets, several discrete attributes strongly affect data characteristics, for example, the labels of observations, which are usually modeled using categorical distributions.
Some of these discretely labeled attributes might be naturally continuous and labeled discretely only because assigning a continuous value to them is cumbersome.
For example, labeling whether a face image contains a beard is relatively easy for humans, while specifying the amount of beard as a continuous attribute is problematic. CelebA (\cite{celeba}) is an example of a dataset of this kind, containing 40 binary labels for face images. Although datasets tend to label attributes like this in discrete ways, many of these attributes may be naturally continuous. So it seems beneficial for the model to find continuous representations of some discretely labeled attributes.

In this paper, to learn insightful representations, we propose a proper structure for the latent space that partitions the representation into two parts: class-related factors and class-independent ones (here, discrete attributes are also called classes). The class-independent part of the representation aims to capture factors of variation shared between different classes, and the class-related part intends to extract factors exclusive to each class. The class-related part can capture the continuous value of a discrete attribute if it is continuous in nature. Since this part should only capture information related to the input class, we utilize an attention mechanism in the latent space to access class-related variation of data.
The prior distribution on the class-independent latent space is modeled as a standard Gaussian distribution (like VAE), while for the class-dependent latent space, we use a mixture of Gaussian as the prior distribution.
Furthermore, to face multimodal data distributions and prevent different modes of the mixture distribution from too much overlapping, we use the Bhattacharyya coefficient in the objective function.

Our framework, which we term \modelname, learns more interpretable and disentangled representations. 
Since completely unsupervised disentanglement learning is challenging (\cite{locatello2019challenging}), similar to \cite{siddharth2017learning} and \cite{revae}, we use a small fraction of labeled data and train our encoder in a semi-supervised manner.
In many downstream tasks, such as classification, it is favorable to be able to put aside those factors of variations that are class-independent (e.g., stroke width or small rotations in the MNIST (\cite{mnist}) dataset) and only focus on class-related features. Our model can be used in such tasks, with a limited number of labeled samples, to provide class-related features for all data points.

Our contributions can be summarized as follows:
\begin{itemize}
    \item We propose the \modelname{} model that jointly models class-related and class-independent factors by choosing a proper structure in the latent space;
    \item we use an attention-like mechanism in VAE's latent space to access class-related factors in an interpretable way;
    \item we introduce appropriate prior distributions for different factors and utilize the Bhattacharyya coefficient to address different modalities of data distribution.
\end{itemize}

\section{RELATED WORK}

Related prior works can be studied in three main aspects: The structure of latent space, supervision, and disentanglement, which are explored followingly.

Initial works on VAEs considered the latent space as a continuous multidimensional space with a standard Gaussian prior. Therefore, these models could not capture either different modalities of data distribution or discrete variations effectively. Nonetheless, several prior works have studied both discrete and continuous variations.
Some models have considered them to be independent (\cite{jointvae, Kim2020SemisupervisedDW}); hence, they cannot extract variations exclusive to particular classes.
Some other models have considered all continuous variations to be class-dependent (\cite{vade, cpvae, AutoEncodingTC}) and so unable to extract variations shared between different classes of data efficiently.
Note that these models can learn shared factors of variation separately for each class using a part of the dataset, which will also waste the model's computation power. Our model, in contrast, learns these factors once using the whole training set, and it also models class-dependent variations.

Recently, \cite{revae} have proposed a semi-supervised model that similarly uses a class-independent set of variables and some class-dependent variables. However, there is a crucial difference between this model and ours. 
Their model calculates class-dependent distribution and then uses this information to find the categorical distribution. In contrast, our work uses the categorical distribution to calculate the class-dependent variables, so we are able to employ an attention mechanism in the latent space and enhance class-related features. Our usage of the Bhattacharyya coefficient is another key difference.
\cite{Antoran2019} also structured the latent space to support both class-related and class-dependent factors. However, their method is completely supervised and hence cannot be used in many cases.
Also, they do not let different modalities of distribution have different variances, so their model does not seem to be applicable when classes are imbalanced. 

Different models can be studied based on how much labeled data they need. It is difficult for many datasets to label all of the samples, while it is challenging to learn meaningful representations with little or no supervision. The existing VAE-based representation learning methods can be categorized into: unsupervised (\cite{betavae, jointvae, factorvae, betatcvae, cpvae}), semi-supervised (\cite{siddharth2017learning, revae, Kim2020SemisupervisedDW}), and supervised ones (\cite{klys2018learning, Antoran2019}).

Learning disentangled representations can be regarded as learning a latent space in which underlying (and meaningful) factors of variation in data can be learned and adjusted separately (\cite{Bengio2013RepresentationLA}). 
An unsupervised and straightforward model to learn disentangled representations is $\beta$-VAE (\cite{betavae}). It simply upweights the $\kl{q_\phi(z|x)}{p(z)}$ part of the vanilla VAE's objective, and is believed to encourage disentanglement of learned representations (\cite{betavae, burgess2018understanding}). There are also several information-theoretic approaches to the disentanglement problem (\cite{klys2018learning, infocatvae, info-vae}). Some models have viewed disentanglement as the independence of marginal distributions of the latent variable. Thus, they have used the notion of Total Correlation ($TC$, \cite{TC}) to introduce alternative objective functions (\cite{factorvae, betatcvae, AutoEncodingTC, Kim2020SemisupervisedDW, Esmaeili2019StructuredDR}). Interestingly, \cite{betatcvae} have shown that the success of $\beta$-VAE in learning disentangled representations can be attributed to penalizing the $TC$ term.

\section{BACKGROUND}

\subsection{Variational Autoencoder}

Variational autoencoder (\cite{vae}) is a latent variable generative model that is capable of performing inference, as well as generation. 
This framework tries to learn the data distribution $p_\theta(x)$ using an empirical distribution $p_D(x)$ on observed data. It assumes an underlying continuous latent variable $z$ and models the data distribution as \mbox{$p_\theta(x) = \int_z p_\theta(x|z) \; p(z) \; dz$}. 
However, the likelihood function ($p_\theta(x) = p(x|\theta)$ denotes the likelihood here) is intractable due to the integration.
Thus, a variational inference distribution $q_\phi(z|x)$ is defined, and the model optimizes a lower bound on the log-likelihood function, called ELBO:
\begin{align*}
    \begin{split}
        \label{eq.elbo}
        \mathcal{L}_\text{ELBO}(x) &= \E{q_\phi(z|x)}{\log p_\theta(x|z)} - \kl{q_\phi(z|x)}{p(z)} \\
    \end{split}
\end{align*}
This lower bound can be maximized in expectation over $p_D(x)$ to obtain the model parameters as 
\begin{equation*}
    \phantom{.}
    \phi^{*}, \theta^{*} = \argmax_{\phi, \theta} \; \E{p_D(x)}{\mathcal{L}_\text{ELBO}(x)}
    .
\end{equation*}

\subsection{Bhattacharyya Coefficient}

The Bhattacharyya coefficient (\cite{bhattacharyya1946measure}) is a symmetric measure of similarity between two probability distributions, defined as
\begin{equation*}
    BC(p_1, p_2) = \int \sqrt{p_1(z)p_2(z)} \; dz.
\end{equation*}
This coefficient is bounded by $0 \leq BC(p_1, p_2) \leq 1$, achieving its maximum when the distributions are equal. 
Note that the Bhattacharyya coefficient can be calculated efficiently for multivariate Gaussian distributions (see the Appendix \ref{app:bhattacharyya} for details).

We have utilized this measure to penalize the overlapping between modalities of distributions. We have observed that doing so leads to a better distinction between classes.

\section{PROPOSED METHOD}

In this section, we present \modelname{} that provides a proper structure in the latent space of VAE. First, the graphical model of the proposed model is introduced. Then, the objective function containing several terms is presented. Finally, the semi-supervised version of our model and an extension, including multiple discrete variables in the latent space, is proposed.

\begin{figure}[h]
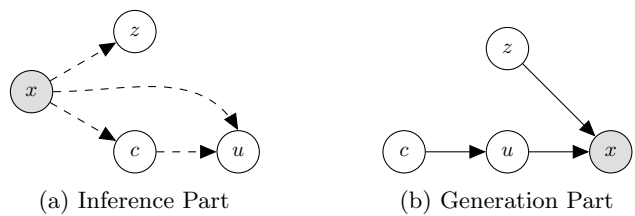

    \centering
    \begin{subfigure}[b]{0.4\columnwidth}
        \centering
        \tikz [scale=0.8, transform shape] {
            \node[obs] (x) {$x$};
            \node[latent,right=of x,yshift=1cm] (z) {$z$};
            \node[latent,right=of x,yshift=-1cm] (c) {$c$};
            \node[latent,right=of c] (u) {$u$};
            \edge[dashed] {x} {z,c}; 
            \edge[dashed] {c} {u};
            \draw [dashed, ->] (x.east) to [out=0,in=120] (u.north);
        }
        \caption{Inference Part}
    \end{subfigure}
    \hfill
    \begin{subfigure}[b]{0.4\columnwidth}
        \centering
        \tikz[scale=0.8, transform shape]{
            \node[latent] (c) {$c$};
            \node[latent,right=of c] (u) {$u$};
            \node[obs,right=of u] (x) {$x$};
            \node[latent,above=of u] (z) {$z$};
            \edge {z, u} {x}; 
            \edge {c} {u};
        }
        \caption{Generation Part}
    \end{subfigure}
    \caption{Model Structure During Inference and Generation}
    \label{fig:graphicalmodel}
\end{figure}

Our method propose a latent space containing 3 parts: $c$, $u$, and $z$. $c$ is a discrete latent variable intended to capture the class of input $x$; $u$ is a continuous latent variable dependent on the value of $c$, which models class-related factors of variation; and $z$ is a continuous latent variable that deals with class-independent factors. $z$ is unaware of the value of $c$ in its computation.
We denote decoder's and encoder's distributions as $p_\theta(.)$ and $q_\phi(.)$, respectively, and show prior latent distributions on $u$ space as $p_\psi(u|c)$. More specifically, $p_\psi(u|c=i)$ is a learned (Gaussian) distribution denoting the $i$-th mode of the prior latent distribution on the $u$ space. Figure~\ref{fig:graphicalmodel} depicts the structure of our framework during inference and generation.

\subsection{Base Objective} \label{subsec:of}

With this structuring of the latent space, the likelihood function can be written as
\begin{equation}
    \label{eq:likelihood}
    \phantom{.} p_\theta(x) = \sum_c \int_u \int_z p_\theta(x|c,u,z) \; p_\psi(u|c) \; p(c) \; p(z) \; dz \; du .
\end{equation}
Because of the integrations in Equation~\ref{eq:likelihood}, optimization of (log) likelihood is intractable. Similar to \cite{vae}, we form a lower bound on the logarithm of the likelihood to address this problem (see Appendix \ref{app:derivation} for details):
\begin{align}
    \begin{split}
        \label{eq.log-likelihood}
        \log p_\theta(x) &\ge \ENB{q_\phi(c|x)} \ENB{q_\phi(u|c,x)} \ENB{q_\phi(z|x)} \left[\log p_\theta(x|c,u,z)\right] \\
        &- \E{q_\phi(c|x)}{\kl{q_\phi(u|c,x)}{p_\psi(u|c)}} \\
        &- \kl{q_\phi(c|x)}{p(c)} \\
        &- \kl{q_\phi(z|x)}{p(z)} \\
        &= \mathcal{L}_1(x)
    \end{split}
\end{align}
Resulting $\mathcal{L}_1(x)$ in Equation \ref{eq.log-likelihood} can be the objective function for a given $x$. However, this function should be averaged over $p_D(x)$ (the true data distribution) to achieve the objective function for the entire empirical distribution.
\begin{equation*}
    \mathcal{L}_1 = \E{p_D(x)}{\mathcal{L}_1(x)}
\end{equation*}

We argue that minimizing
\begin{equation*}
    \E{p_D(x)}{\kl{q_\phi(c|x)}{p(c)}}
\end{equation*}
does not necessarily lead the model to the desired state. In the extreme case, the KL divergence value can tend to zero which means that for all $x$ that have positive value of $p_D(x)$, $q_\phi(c|x)$ is nearly $p(c)$, so $x$ and $c$ tend to be independent.
(This argument holds for KL divergences of $u$ and $z$ too, but this problem is addressed differently in Subsection~\ref{subsec:capacity}.)
To address the aforementioned issue, we define
\begin{align*}
    q_\phi(c) = \E{p_D(x)}{q_\phi(c|x)}
\end{align*}
as \emph{aggregate discrete posterior distribution}, similar to aggregate posterior distribution by \cite{aae}, and rewrite the third term of $\mathcal{L}_1(x)$ in Equation~\ref{eq.log-likelihood}, averaged over $p_D(x)$ as (see Appendix \ref{app:disc-kl} for proof)
\begin{align}
    \label{eq:discrete-kl}
    \begin{split}
        \phantom{.}
        &\ENB{p_D(x)} [\kl{q_\phi(c|x)}{p(c)}] = \\
        &= \underbrace{\ENB{p_D(x)} \left[\kl{q_\phi(c|x)}{q_\phi(c)}\right]}_\textrm{$A$} + \underbrace{\kl{q_\phi(c)}{p(c)}}_\textrm{$B$}
        .
    \end{split}
\end{align}

Minimizing the term $B$ of this equation is favorable since it makes the aggregate discrete posterior distribution close to the prior. On the other hand, minimizing the term $A$ seems undesirable since we do not want $q_\phi(c|x)$ to be close to the prior for all $x$. In fact, a good model should predict the value of $c$ given $x$ with high certainty. To this end, we suggest replacing the $A$ term with an entropy term: $H(q_\phi(c|x))$. This way, a trade-off occurs: the model tends to be almost sure about the value of $c$ related to a specific $x$, and the term $B$ ensures that overall $q_\phi(c)$ tends to $p(c)$.

So, up until now, the objective function is formulated as follows:
\begin{align}
    \begin{split}
        \label{eq.obj-no-bc}
        &\mathcal{L}_2 = \\
        &\quad \ENB{p_D(x)} \ENB{q_\phi(c|x)} \ENB{q_\phi(u|c,x)} \ENB{q_\phi(z|x)} \left[\log p_\theta(x|c,u,z)\right] \\
        &\quad - \ENB{p_D(x)} \E{q_\phi(c|x)}{\kl{q_\phi(u|c,x)}{p_\psi(u|c)}} \\
        &\quad - \ENB{p_D(x)} \left[H(q_\phi(c|x))\right] \\
        &\quad - \kl{q_\phi(c)}{p(c)} \\
        &\quad - \E{p_D(x)}{\kl{q_\phi(z|x)}{p(z)}}
    \end{split}
\end{align}

\subsection{Penalizing Overlapping Class Distributions} \label{subsec:bc}

As stated, $p_\psi(u|c)$ denote different prior distributions on $u$ space, one for each value of $c$, which are being trained by the objective function in Equation \ref{eq.obj-no-bc}. We observed that optimizing this objective function leads these distributions to have large intersections with one another.
Since $p_\psi(u|c)$ for different values of $c$ are meant to model different modalities of data distribution, with respect to the attribute represented by $c$, the intersection of $p_\psi(u|c=i)$ and $p_\psi(u|c=j)$ (for $i \neq j$) should be low.
To this end, we suggest using a Bhattacharyya coefficient term to be minimized in the objective function. Depending on the factor that $c$ represents, a small amount of intersection might be tolerable (or desirable). For example, when we are modeling the \textit{Hair Color} attribute of a face image, \textit{Black} and \textit{Brown} should be two of the modalities of the distribution, and there are hair colors that can be classified as both black and brown, somewhere in between. So for an attribute like this, it is logical to let the $p_\psi(u|c=i)$ distributions have some amount of intersection. This is why we used a threshold to penalize the model only if the Bhattacharyya coefficient's value is higher than that of the threshold.

The objective function, considering the new Bhattacharyya coefficient term, is
\begin{align}
    \begin{split}
        \label{eq.obj-bc}
        \phantom{,} \mathcal{L}_3 &= \mathcal{L}_2 - \sum_{i=1}^{L} \sum_{j=i}^{L} BC(p_\psi(u|c=i), p_\psi(u|c=j))
        ,
    \end{split}
\end{align}
in which $BC(., .)$ denotes the Bhattacharyya coefficient, and $L$ is the number of different values that $c$ can take.

\subsection{Controlled Capacity Increase} \label{subsec:capacity}

\cite{factorvae, jointvae} show that \mbox{$\E{p_D(x)} {\kl{q_\phi(z|x)}{p(z)}}$}, the last term in Equation \ref{eq.obj-no-bc}, is an upper bound on the mutual information between $x$ and $z$
(see Appendix \ref{app:cont-kl} for proof):
\begin{equation*}
    \E{p_D(x)}{\kl{q_\phi(z|x)}{p(z)}}\geq I_{q_\phi}(x; z)
\end{equation*}
Similarly, we can prove that the term \mbox{$\ENB{p_D(x)}\E{q_\phi(c|x)}{\kl{q_\phi(u|c,x)}{p_\psi(u|c)}}$} in Equation~\ref{eq.obj-no-bc} is an upper bound on the expected mutual information between $x$ and $u$ given $c$
(see Appendix \ref{app:cont-kl} for proof):
\begin{multline*}
    \ENB{p_D(x)} \E{q_\phi(c|x)}{\kl{q_\phi(u|c,x)}{p_\psi(u|c)}} \\
    \geq \E{q_\phi(c)}{I_{q_\phi}(x; u | c)}
\end{multline*}
Although minimizing these KL divergences may benefit disentanglement, it simultaneously degrades reconstruction quality because they are upper bounds on mutual information between latent variables and $x$. \cite{burgess2018understanding} propose to control and gradually increase these KL divergences during training so that the upper bounds on mutual information will gently increase. 
Hence, the objective in Equation \ref{eq.obj-bc} will turn to the following, which is the final objective of our model:
\begin{align}
    \begin{split}
        \label{eq.obj-capacity}
        &\mathcal{L}_\text{\modelname} = \\
        & \ENB{p_D(x)} \ENB{q_\phi(c|x)} \ENB{q_\phi(u|c,x)} \ENB{q_\phi(z|x)} \left[\log p_\theta(x|c,u,z)\right] \\
        & - \gamma_u \; \E{p_D(x)}{\left| \E{q_\phi(c|x)}{\kl{q_\phi(u|c,x)}{p_\psi(u|c)}} - C_u \right|} \\
        & - \gamma_h \; \ENB{p_D(x)} \left[H(q_\phi(c|x))\right] \\
        & - \gamma_c \; \kl{q_\phi(c)}{p(c)} \\
        & - \gamma_z \; \E{p_D(x)}{|\kl{q_\phi(z|x)}{p(z)} - C_z|} \\
        & - \gamma_{bc} \; \sum_{i=1}^{L} \sum_{j=i}^{L} \max(BC(p_\psi(u|c=i), p_\psi(u|c=j)) - \delta, 0)
    \end{split}
\end{align}
In this equation, $C_z$ and $C_u$ are the information capacities and are gradually increased during training.
We have observed that this technique leads to a gradual decrease in $q_\phi(z|x)$ and $q_\phi(u|c,x)$ variances and makes the model more sure about these distributions.
$\gamma$ variables are hyperparameters of the model, which $\gamma_c$, $\gamma_u$, and $\gamma_z$ are usually set to the same value.
$\delta$ denotes the amount of intersection that is tolerable (see Subsection~\ref{subsec:bc}).

\subsection{Training Process}

We model $q_\phi(c,u,z|x)$ and $p_\theta(x|c,u,z)$ distributions as deep convolutional neural networks (the encoder and decoder network, respectively).
$q_\phi(c|x)$ is assumed to be a categorical distribution. Since we require sampling from this discrete distribution,
a Gumbel-softmax distribution (\cite{maddison-gumbel-softmax, jang-gumbel-softmax}) is used to allow gradient backpropagation.

\begin{figure}[th]
    \centering
    \includegraphics[width=\columnwidth]{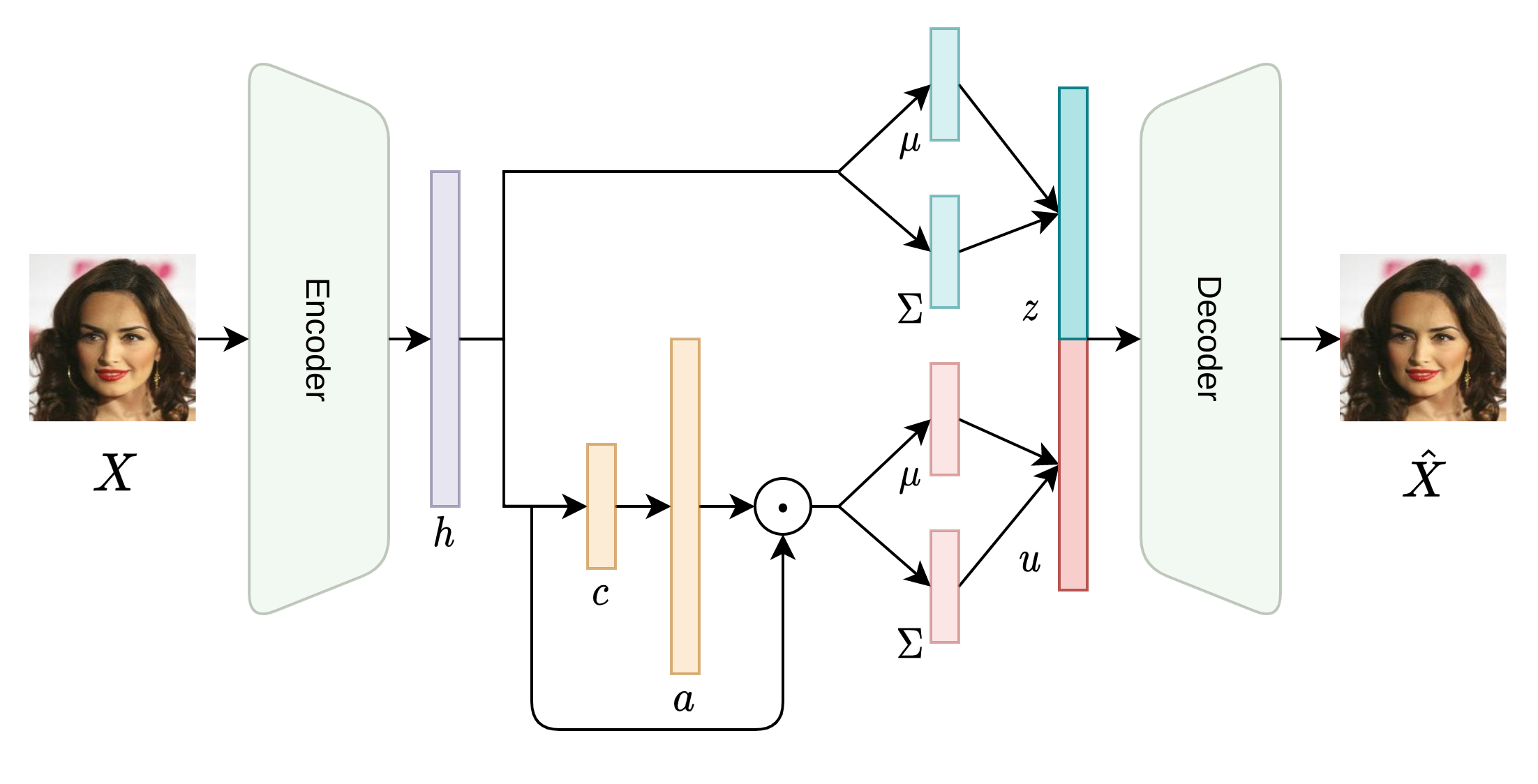}
    \caption{Forward Flow During Training.
    First, input $x$ is fed to the encoder network to obtain a feature vector called $h$. Logits of the $q_\phi(c|x)$ distribution are found by a linear layer on $h$. We take a sample $c_0$ from this distribution using the Gumbel-softmax trick.
    For finding the required $\mu$ and $\Sigma$ of $q_\phi(u|c,x)$ in the forward pass, an attention mechanism is utilized.
    $c_0$ is passed through a linear layer, followed by a Sigmoid, to form the attention map, $a$.
    $a$, which is of the same size as $h$, shows the model how important every entry of $h$ is with respect to $c_0$. $h \cdot a$ is considered to be the $c_0$-related feature vector. The $\mu$ and $\Sigma$ of $q_\phi(u|c,x)$ are found by a linear layer on $h \cdot a$ so that they only contain features related to $c_0$. On the other track, parameters of $q_\phi(z|x)$ distribution are found by a linear layer using only $h$. At last, concatenated samples of $q_\phi(z|x)$ and $q_\phi(u|c,x)$ are fed into the decoder.
    }
    \label{fig:flow}
\end{figure}

Calculating $q_\phi(c)$ is a time-consuming task, so we estimate it using the current mini-batch.
Both $q_\phi(u|c,x)$ and $q_\phi(z|x)$ are modeled as Gaussian distributions and hence can be sampled using the reparameterization trick (\cite{vae}).
Another set of parameters in the model ($\psi$) are the parameters of prior distributions on the latent space $u$. We assume $p_\psi(u|c=i)$ to be a Gaussian distribution with the learnable parameters $\mu$ and diagonal $\Sigma$, for every possible value of $i$.
$p(c)$ is the categorical random variable's prior distribution, and $p(z)$ is a standard Gaussian distribution. Figure~\ref{fig:flow} shows the flow of our framework during training. 
We utilize an attention mechanism that helps the model know where to attend while capturing class-related attributes.
During generation, first $c_0$ is sampled from $p(c)$, then a sample from $p_\psi(u|c=c_0)$ and a sample from $p(z)$ are concatenated together and the result is fed into the decoder. More details of the architecture and also the training hyperparameters are presented in the Appendices \ref{app:arch} and \ref{app:training}.

\subsection{Semi-Supervised Learning}

Besides training the model with the objective in Equation \ref{eq.obj-capacity}, we optimized the encoder network in a semi-supervised fashion. This is because our model depends heavily on the $q_\phi(c|x)$ distribution to be accurate. To this end, we use a simple cross-entropy loss
\begin{equation}  
    \phantom{,}\mathcal{L}_{S} = -\E{p_S(x)}{
        \sum_{i=1}^{L} \mathbbm{1}\{y=i\} \log q_\phi(c=i|x)
    },
\end{equation}
in which $p_S(x)$ is the labeled data distribution, $y$ is the true label of $x$, and $L$ is the number of different values that $y$ or $c$ can take.
$\mathcal{L_\text{\modelname}}$ and $\mathcal{L}_{S}$ are optimized in turn.
The former affects all of the model's parameters ($\phi$, $\theta$, $\psi$), and the latter only affects the encoder's ($\phi$).

\subsection{Generalization of \texorpdfstring{$c$}{c} and \texorpdfstring{$u$}{u}}

\begin{figure}
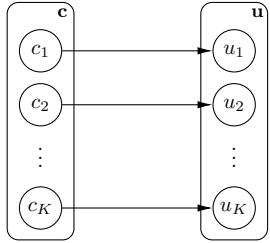

    \centering
    \tikz[scale=0.8, transform shape]{
        \node[latent, label={[label distance=1mm]70:$\Vec{c}$}] (c1) {$c_1$};
        \node[latent, below=of c1, yshift=0.82cm] (c2) {$c_2$};
        \node[latent, below=of c2] (ck) {$c_K$};
        \node at ($(c2)!.45!(ck)$) {\vdots};
        \plate [inner sep=.15cm,yshift=.2cm, rounded corners=1ex] {plate1} {(c1)(c2)(ck)}{};
        
        \node[label={[label distance=1mm]70:$\Vec{u}$}, latent, right=of c1, xshift=1.5cm] (u1) {$u_1$};
        \node[latent, below=of u1, yshift=0.82cm] (u2) {$u_2$};
        \node[latent, below=of u2] (uk) {$u_K$};
        \node at ($(u2)!.45!(uk)$) {\vdots};
        \plate [inner sep=.15cm,yshift=.2cm, rounded corners=1ex] {plate2} {(u1)(u2)(uk)}{};
        
        \edge[-{Latex[length=2mm,width=1mm]}] {c1} {u1}
        \edge[-{Latex[length=2mm,width=1mm]}] {c2} {u2}
        \edge[-{Latex[length=2mm,width=1mm]}] {ck} {uk}
    }
    \caption{The Model's Latent Structure in a Multi-Label Setting}
    \label{fig:generalization}
\end{figure}

Up until now, for the sake of simplicity, we assumed a single $c$ variable (with multiple possible values) and a single $u$ vector, dependent on $c$. The model can be generalized to a multi-label setting as
\begin{gather*}
    \Vec{c} = \{c_1, c_2, \dots, c_K\} \\
    \phantom{,}\Vec{u} = \{u_1, u_2, \dots, u_K\},
\end{gather*}
in which every $u_i$ is dependent on $c_i$ and is independent of any $c_j, \; j \neq i$. The graphical model of this setting is shown in Figure~\ref{fig:generalization}.
In this case, every $c_i$ can have $L_i$ possible different values, and we can write all the model's equations accordingly
(see Appendix \ref{app:generalization} for details).

\section{EXPERIMENTS}

In this section, we first evaluate our model's performance quantitatively by discussing disentanglement scores in an unsupervised and semi-supervised fashion. We also show that our objective function can help downstream tasks, such as classification.
We then present some qualitative results on MNIST and CelebA datasets.
Furthermore, we discuss the Bhattacharyya coefficient's role (Subsection~\ref{subsec:bc}) in the quality of learned latent space.

\subsection{Quantitative Evaluation}

Measuring disentanglement requires having a dataset about which we know ground truth factors of variation. \cite{dsprites} introduced the dSprites dataset, a dataset of 2D shapes with six independent latent factors. This dataset is usually used to measure disentanglement scores. \cite{betavae} proposed a metric for quantifying disentanglement that is the accuracy of a linear classifier. However, this metric is sensitive to hyperparameters and has a failure mode (\cite{factorvae}). Another metric that we refer to as the Factor score is proposed by \cite{factorvae} to address these weaknesses. This metric uses a majority-vote classifier and does not require optimization.
Table~\ref{table.unsupervised-dsprites} shows the performance of some purely unsupervised methods, namely
Vanilla VAE (\cite{vae}), $\beta$-VAE (\cite{betavae}), FactorVAE (\cite{factorvae}), $\beta$-TCVAE (\cite{betatcvae}), HFVAE (\cite{Esmaeili2019StructuredDR}), Guided-VAE, and Guided-$\beta$-TCVAE (\cite{guided}) in addition to unsupervised version of our model.
Our model achieves a Factor score of
$0.775$ $(\pm 0.02)$
over five different runs.
None of these models, including ours, has captured the \textit{Shape} factor of this dataset in a disentangled way. This phenomenon implies that a little amount of supervision might be needed. As our framework's primary goal is to learn factors in a semi-supervised fashion, we have also measured its performance in semi-supervised settings with different supervision amounts.
Table~\ref{table:supervised-dsprites} shows our model's Factor score given different amounts of supervision.
We have only accessed the value of \textit{Shape} factor and let the other factors be captured automatically through $z$ variable.
Observations show that when there are too few labeled samples, the performance is worse than the unsupervised approach and these samples jeopardize the Factor score. This is because the model faces those few labeled samples very often, and it overfits. On the other hand, with a reasonable amount of supervision, the model can achieve high Factor score values.

\begin{table}[h]
    \caption{Unsupervised Disentanglement Score.
    Factor score (higher is better) of multiple unsupervised methods over the dSprites dataset. We assumed $c$ to be a 3-state discrete variable and $d_u = 1$. We also assumed $d_z = 5$ so that the latent dimension of our model is 6. Other methods' number of latent dimensions is 6, too.
    Data are partially obtained from \cite{guided}.}
    \label{table.unsupervised-dsprites}
    \begin{center}
        \begin{tabular}{lc}
            \textbf{Model}    &\textbf{Factor Score} \\
            \hline
            VAE                  & 0.41 \\
            $\beta$-VAE ($\beta = 2$)      & 0.58 \\
            FactorVAE ($\gamma = 35$)      & 0.71 \\
            $\beta$-TCVAE ($\alpha = 1, \beta = 5, \gamma = 1$) & 0.70  \\
            HFVAE                & 0.63 \\
            Guided-VAE           & 0.67 \\
            Guided-$\beta$-TCVAE & 0.73 \\
            \hline
            \modelname{} (Ours)    & \textbf{0.77}
        \end{tabular}
    \end{center}
\end{table}

\begin{table}[h]
    \caption{Semi-Supervised Disentanglement Score.
    Factor score (higher is better) of our method over the dSprites dataset in a semi-supervised fashion, with different amounts of labeled data (on each of these settings, our model is run five times). Like the unsupervised case, $c$ is a 3-state discrete variable, $d_u = 1$, and $d_z = 5$. We only used \textit{Shape} labels to be captured in $c$.
    }
    \label{table:supervised-dsprites}
    \begin{center}
        \begin{tabular}{ccc}
            \textbf{\# Labeled Data} &\textbf{Percentage}    &\textbf{Factor Score} \\
            \hline
            100      & 0.013\%    & 0.626 ($\pm$ 0.04) \\
            737      & 0.1\%      & 0.718 ($\pm$ 0.03) \\
            1000     & 0.13\%     & 0.735 ($\pm$ 0.04) \\
            3686     & 0.5\%      & 0.881 ($\pm$ 0.11) \\
            10000    & 1.35\%     & 0.905 ($\pm$ 0.10) \\
        \end{tabular}
    \end{center}
\end{table}

Furthermore, our model is able to perform well on classification tasks. To show this, we trained our model on the MNIST dataset, with the label values of 256 images for semi-supervision. This model's encoder can be seen as a classifier, and it achieves a classification accuracy of $95\%$ on the test dataset. Additionally, we trained a baseline model using the same architecture by only incorporating the same 256 labeled samples. In other words, we used the same architecture to train a classifier using 256 labeled images. In this case, the best accuracy we were able to achieve was $83\%$. This simple experiment confirms that our latent space structure and objective function can efficiently use unlabeled data to enhance its accuracy. The $95\%$ result is comparable with the results of \cite{revae}, another related semi-supervised model. Nonetheless, our model is not primarily designed for classification tasks. This experiment aimed to show that our model can learn discrete variables effectively and preserve disentanglement simultaneously.

\subsection{Qualitative Results}

\begin{figure}[h]
    \centering
    \begin{subfigure}[b]{0.45\columnwidth}
        \centering
        \includegraphics[width=1\columnwidth]{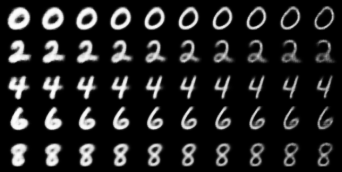}
        \caption{Thickness Factor}
    \end{subfigure}
    \begin{subfigure}[b]{0.45\columnwidth}
        \centering
        \includegraphics[width=1\columnwidth]{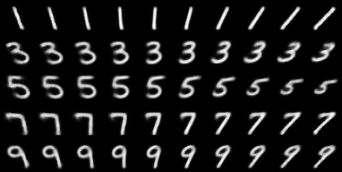}
        \caption{Angle Factor}
    \end{subfigure}
    \begin{subfigure}[b]{0.65\columnwidth}
        \centering
        \includegraphics[width=1\columnwidth]{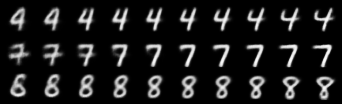}
        \caption{A Number of Class-Related Factors}
    \end{subfigure}
    \caption{Class-Related and Class-Independent Factors Learned on the MNIST Dataset}
    \label{fig:mnist_lantent}
\end{figure}

The disentanglement results of our model on the MNIST dataset are presented in Figure~\ref{fig:mnist_lantent}. We have used latent traversals to show that our model has successfully discovered and disentangled the digits' angle and thickness as class-independent factors of variation. It has also discovered two different writing styles of the digit 4, the middle line of the digit 7, and the relative size of circles in the digit 8 as examples of class-related variation. 

\begin{figure}[h]
    \centering
    \includegraphics[width=0.6\columnwidth]{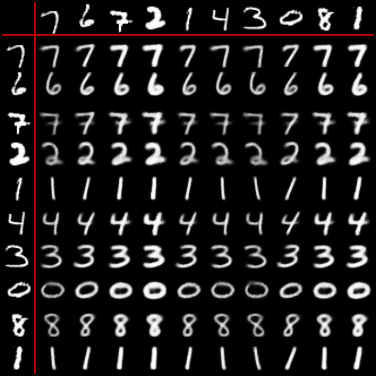}
    \caption{Attributes (apart from the digit number) of the first row images have been transferred to the images of the leftmost column.}
    \label{fig:mnist-attribute-transfer}
\end{figure}

As another application of our framework, we can transfer general attributes from one sample to another instance. In the attribute-transfer procedure, the discrete latent variables and the class-related variables are preserved, and the class-independent variables are transferred. An example of attribute transfer on the MNIST dataset is shown in Figure~\ref{fig:mnist-attribute-transfer}.

\begin{figure*}[t]
    \centering
    \begin{subfigure}[t]{\columnwidth}
        \centering
        \includegraphics[width=\columnwidth]{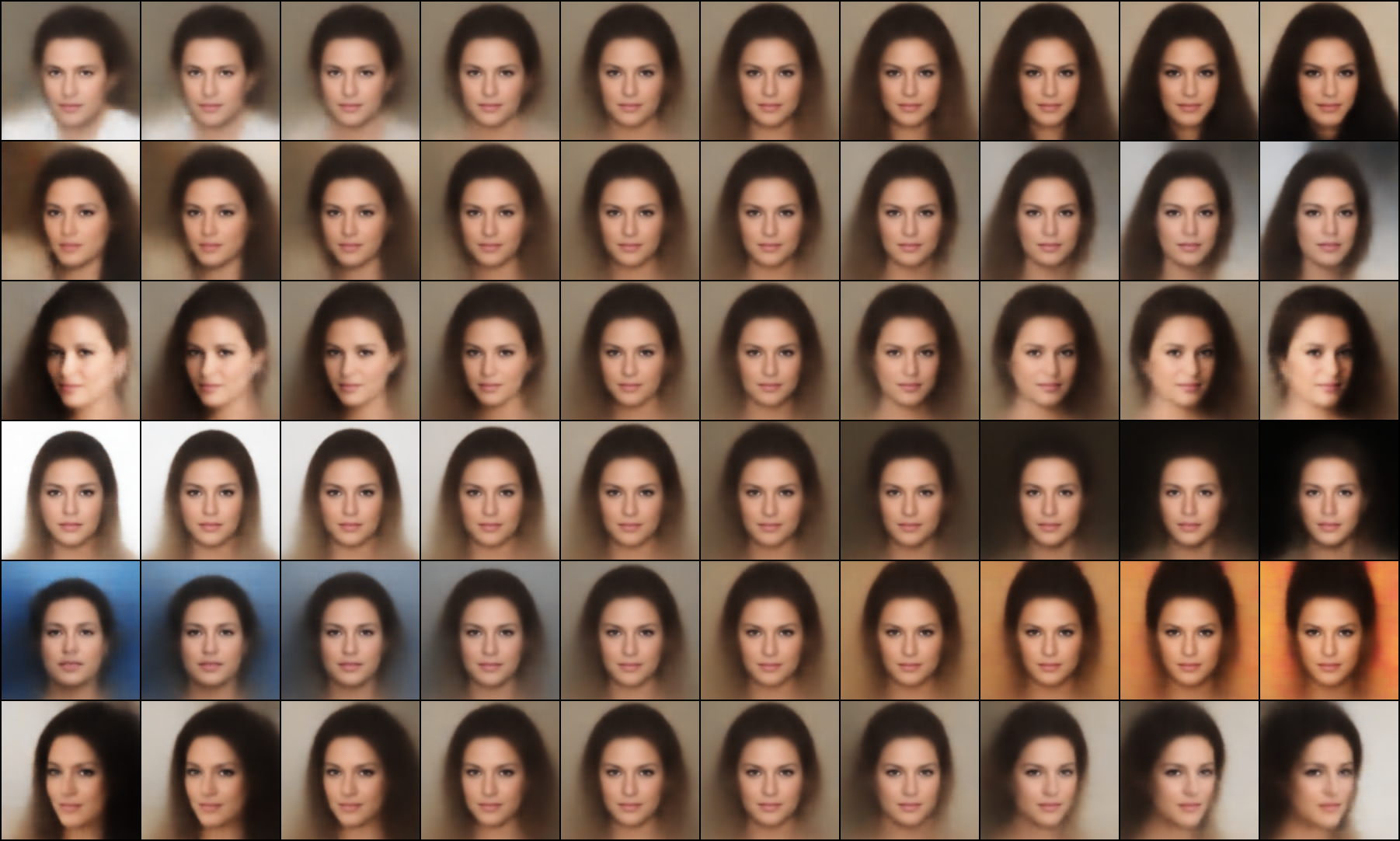}
        \caption{Traversal on Dimensions of $z$. Our model has been able to capture \textit{Azimuth}, \textit{Background Color}, \textit{Background Illumination}, \textit{Hair Length}, and other class-independent factors in the $z$ variable.}
    \end{subfigure}
    \hfill
    \begin{subfigure}[t]{\columnwidth}
        \centering
        \includegraphics[width=\columnwidth]{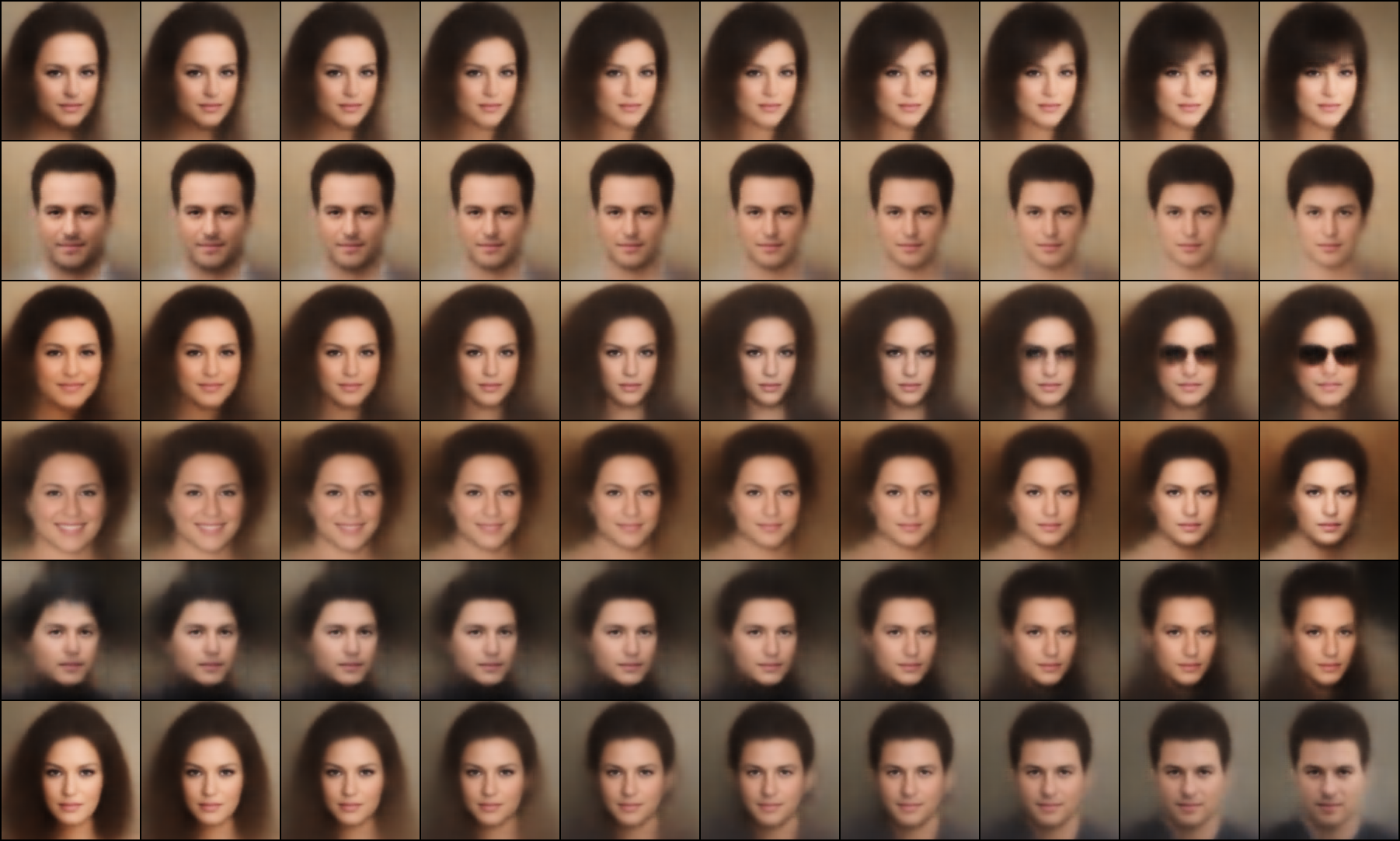}
        \caption{Traversal on Dimensions of $u$. Rows denote \textit{Bang}, \textit{Receding Hairline}, \textit{Eyeglasses}, \textit{Smiling}, \textit{Hat}, and \textit{Gender}, respectively.}
    \end{subfigure}
    \caption{CelebA Latent Space Traversals}
    \label{fig:celeba-traversal}
\end{figure*}

We further investigate our model's semi-supervised performance on the CelebA dataset, with 0.5\% (Similar to \cite{anandkumar}) of training data as labeled instances. Figure~\ref{fig:celeba-traversal} shows this experiment's latent space traversal. Our model has captured many class-independent attributes without supervision and some class-related attributes in a semi-supervised fashion.

\begin{figure*}[]
    \centering
    \begin{subfigure}{\textwidth}
        \centering
        \label{fig:celeba-hair}
        \includegraphics[width=\textwidth]{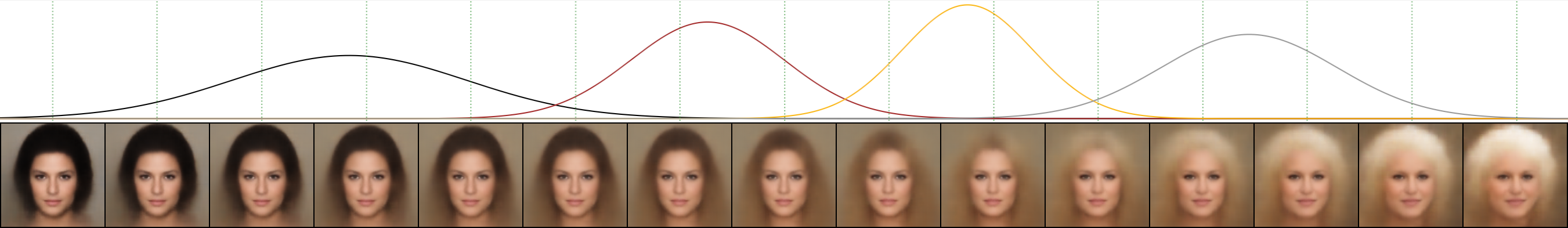}
        \caption{Traversal of \textit{Hair Color}. Modes represent Black, Brown, Blond, and Gray/White, from left to right}
    \end{subfigure}
    \begin{subfigure}{\textwidth}
        \centering
        \label{fig:celeba-skin}
        \includegraphics[width=\textwidth]{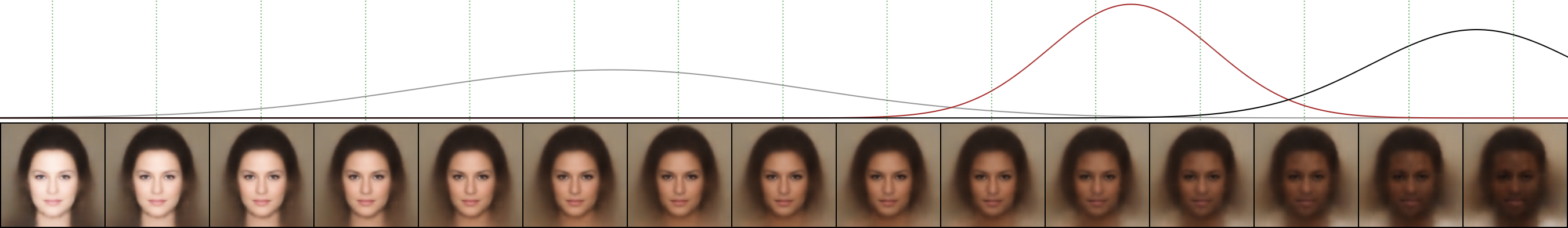}
        \caption{Traversal of \textit{Skin Tone}. Modes represent White, Brown, and Black, from left to right}
    \end{subfigure}
    \caption{Traversals of \textit{Hair Color} and \textit{Skin Tone} Factors of Faces. In each subfigure, the learned distributions on latent space are depicted above. 
    We have multiple generated images under the distributions' plot that their corresponding element of $u$ is set to the value indicated by the green, dotted, vertical line above it. Traversing these latent dimensions results in a smooth traversal of \textit{Hair Color} and \textit{Skin Tone} factors.}
    \label{fig:celeba-hair-skin}
\end{figure*}

To better illustrate the way our framework models multimodality of data distribution, Figure~\ref{fig:celeba-hair-skin} shows prior distributions of \textit{Hair Color} and \textit{Skin Tone} attributes, alongside with multiple images generated corresponding to different samples of priors. 

\subsection{The Effect of Bhattacharyya Coefficient} \label{subsec:eff-bc}

\begin{figure}[]
    \centering
    \includegraphics[width=\columnwidth]{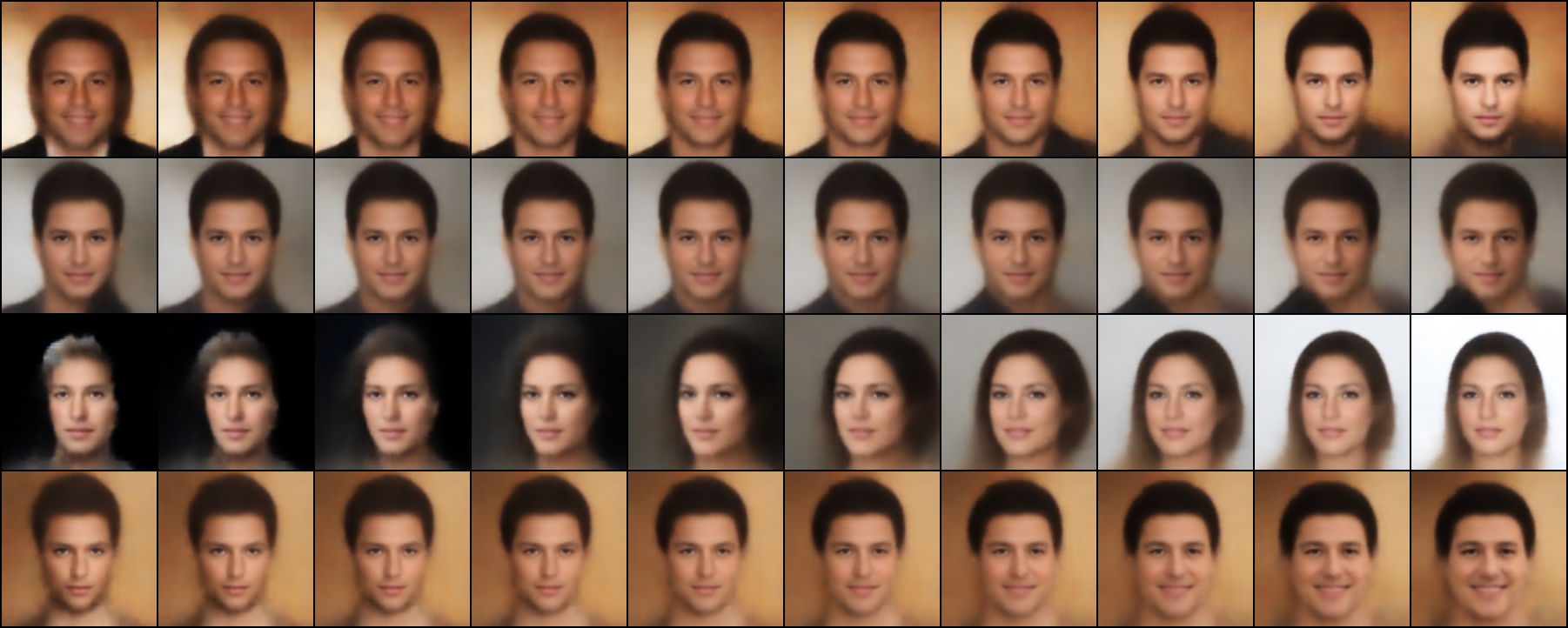}
    \caption{Traversal on Dimensions of $u$ Without BC. Rows were meant to capture \textit{Skin Tone}, \textit{Beard}, \textit{Eyeglasses}, and \textit{Hat} factors, from top to bottom.}
    \label{fig:celeba-no-bc-u}
\end{figure}

We utilized the Bhattacharyya coefficient to penalize overlapping class distributions. To find about the effect of this new term, we trained a semi-supervised model on the dSprites dataset, same as the one that used 0.5\% of labeled data in Table~\ref{table:supervised-dsprites}, but without using the Bhattacharyya coefficient. We observed that this could damage the model's performance on Factor score. More specifically, the model's Factor score dropped from
$0.881$
to
$0.829$. We further analyzed the Bhattacharyya coefficient effect on the CelebA sample generation.
Figure~\ref{fig:celeba-no-bc-u} illustrates $u$ traversals for a model without BC term, in which many factors of variation are not captured (also see Appendix \ref{app:eff-bc}).

\section{CONCLUSION}

We have proposed \modelname, a method for learning disentangled representations, which considers different modalities of the data distribution and split the representation into class-related and class-independent parts. Hence, our model is semantically appealing and is able to utilize data more efficiently by using the whole dataset for learning class independent variables. We have also modified our model's objective function to achieve a more apparent distinction between different modalities of data distribution and better disentanglement results. Furthermore, we have evaluated our model's performance on the dSprites, MNIST, and CelebA datasets quantitatively and qualitatively.
In the future, we would like to incorporate information-theoretic approaches, particularly the use of Total Correlation, into our method.

\bibliography{reference}

\newpage
\appendix
\onecolumn

\section{PROOFS}

In this section, we present proofs and details of the equations that have appeared in the paper.

\subsection{Calculating Bhattacharyya Coefficient} \label{app:bhattacharyya}

Generally, the Bhattacharyya distance between two distribution is defined as
\begin{equation*}
    \phantom{.}D_B(p_1, p_2) = - \ln BC(p_1, p_2),
\end{equation*}
where $BC(., .)$ denotes the Bhattacharyya coefficient.

The Bhattacharyya distance between two multivariate Gaussian distributions $p_i = \Norm(\mu_i, \Sigma_i)$, can be calculated efficiently using
\begin{equation*}
    \phantom{,}D_B(p_1, p_2) = \frac{1}{8}(\mu_1 - \mu_2)^{T}\Sigma^{-1}(\mu_1 - \mu_2) + \frac{1}{2}\ln(\frac{\det \Sigma}{\sqrt{\det\Sigma_1 \det \Sigma_2}}), 
\end{equation*}
in which $\Sigma = \frac{\Sigma_1 + \Sigma_2}{2}$.

\subsection{Lower Bound of Log Likelihood} \label{app:derivation}

Considering the likelihood function in Equation~\ref{eq:likelihood}, we can derive the base objective function in Equation~\ref{eq.log-likelihood} in the following manner. In the fourth line, Jensen's inequality is used.
\begin{align*}
    \begin{split}
        \log p_\theta(x) &= \log \sum_c \int_u \int_z
        p_\theta(x|c,u,z) \; p_\psi(u|c) \; p(c) \; p(z) \; 
        \frac{q_\phi(c,u,z|x)}{q_\phi(c,u,z|x)} \;
        dz \; du \\
        &= \log \sum_c \int_u \int_z q_\phi(c,u,z|x) \;
        p_\theta(x|c,u,z) \:
        \frac{p_\psi(u|c)}{q_\phi(u|c,x)} \:
        \frac{p(c)}{q_\phi(c|x)} \:
        \frac{p(z)}{q_\phi(z|x)} \\
        &= \log \ENB{q_\phi(c|x)} \ENB{q_\phi(u|c,x)} \E{q_\phi(z|x)}{p_\theta(x|c,u,z) \:
        \frac{p_\psi(u|c)}{q_\phi(u|c,x)} \:
        \frac{p(c)}{q_\phi(c|x)} \:
        \frac{p(z)}{q_\phi(z|x)}} \\
        &\ge \ENB{q_\phi(c|x)} \ENB{q_\phi(u|c,x)} \E{q_\phi(z|x)}{\log p_\theta(x|c,u,z) +
        \log \frac{p_\psi(u|c)}{q_\phi(u|c,x)} +
        \log \frac{p(c)}{q_\phi(c|x)} +
        \log \frac{p(z)}{q_\phi(z|x)}} \\
        &= \ENB{q_\phi(c|x)} \ENB{q_\phi(u|c,x)} \ENB{q_\phi(z|x)} \left[\log p_\theta(x|c,u,z)\right] \\
        &\qquad - \E{q_\phi(c|x)}{\kl{q_\phi(u|c,x)}{p_\psi(u|c)}}
        - \kl{q_\phi(c|x)}{p(c)}
        - \kl{q_\phi(z|x)}{p(z)}
    \end{split}
\end{align*}

\subsection{Expectation of Discrete KL Divergence} \label{app:disc-kl}

The proof of Equation~\ref{eq:discrete-kl} can be written as follows. The Shannon entropy of $q_\phi(c)$ is added and subtracted in the first line.
\begin{align*}
    \begin{split}
        &\ENB{p_D(x)} [\kl{q_\phi(c|x)}{p(c)}] = 
        \ENB{p_D(x)} \ENB{q_\phi(c|x)} \left[\log \frac{q_\phi(c|x)}{p(c)}\right] + H(q_\phi(c)) - H(q_\phi(c)) \\
        & \qquad = \ENB{p_D(x)} \ENB{q_\phi(c|x)} \left[\log q_\phi(c|x)\right] + 
        \underbrace{\E{q_\phi(c)}{\log \frac{1}{q_\phi(c)}}}_\textrm{$H(q_\phi(c))$}
        - \ENB{p_D(x)} \left[\sum_c q_\phi(c|x) \log p(c)\right] - H(q_\phi(c)) \\
        & \qquad = \ENB{p_D(x)} \ENB{q_\phi(c|x)} \left[\log q_\phi(c|x) + \log \frac{1}{q_\phi(c)}\right] - \sum_c \log p(c) \; \underbrace{\ENB{p_D(x)} \left[q_\phi(c|x)\right]}_\textrm{$q_\phi(c)$} - \sum_c q_\phi(c) \log \frac{1}{q_\phi(c)} \\
        & \qquad = \ENB{p_D(x)} \E{q_\phi(c|x)}{\log \frac{q_\phi(c|x)}{q_\phi(c)}} -\sum_c q_\phi(c) \log \frac{p(c)}{q_\phi(c)} \\
        & \qquad = \ENB{p_D(x)} \left[\kl{q_\phi(c|x)}{q_\phi(c)}\right] + \kl{q_\phi(c)}{p(c)}
    \end{split}
\end{align*}

\subsection{Relationship between Mutual Information and Expected Continuous KL Divergences} \label{app:cont-kl}

Regarding to Section~\ref{subsec:capacity}, if we define $q_\phi(z) = \E{p_D(x)}{q_\phi(z|x)}$, we can write
\begin{align*}
    \begin{split}
        \E{p_D(x)}{\kl{q_\phi(z|x)}{p(z)}} &= \ENB{p_D(x)} \E{q_\phi(z|x)}{\log \frac{q_\phi(z|x)}{p(z)}} \\
        &= \E{q_\phi(z,x)}{\log \left(\frac{q_\phi(z|x)}{p(z)} \: \frac{q_\phi(z)}{q_\phi(z)} \right)} \\
        &= \E{q_\phi(z,x)}{\log \frac{q_\phi(z|x)}{q_\phi(z)}} + \E{q_\phi(z,x)}{\log \frac{q_\phi(z)}{p(z)}} \\
        &= \E{q_\phi(z,x)}{\log \frac{q_\phi(z,x)}{p_D(x) q_\phi(z)}} + \E{q_\phi(z)}{\log \frac{q_\phi(z)}{p(z)}} \\
        &= I_{q_\phi}(x; z) + \kl{q_\phi(z)}{p(z)} \\
        &\geq I_{q_\phi}(x; z)
        .
    \end{split}
\end{align*}

So $\E{p_D(x)}{\kl{q_\phi(z|x)}{p(z)}}$ is an upper bound on $I_{q_\phi}(x; z)$ (\cite{jointvae, factorvae}).

We can also define $q_\phi(u|c) = \E{p_D(x)}{q_\phi(u|c,x)}$ and write
\begin{align*}
    \begin{split}
        \ENB{p_D(x)}\E{q_\phi(c|x)}{\kl{q_\phi(u|c,x)}{p_\psi(u|c)}} &= \ENB{p_D(x)}\ENB{q_\phi(c|x)}\E{q_\phi(u|x,c)}{\log \frac{q_\phi(u|c,x)}{p_\psi(u|c)}} \\
        &= \E{q_\phi(c,u,x)}{\log \left(\frac{q_\phi(u|c,x)}{p_\psi(u|c)} \: \frac{q_\phi(u|c)}{q_\phi(u|c)} \right)} \\
        &= \E{q_\phi(c,u,x)}{\log \frac{q_\phi(u|c,x)}{q_\phi(u|c)}} +
        \E{q_\phi(c,u,x)}{\log \frac{q_\phi(u|c)}{p_\psi(u|c)}} \\
        &= \ENB{q_\phi(c)}\E{q_\phi(u,x|c)}{\log \frac{q_\phi(u,x|c)}{q_\phi(x|c) q_\phi(u|c)}} +
        \E{q_\phi(c,u)}{\log \frac{q_\phi(u|c)}{p_\psi(u|c)}} \\
        &= \E{q_\phi(c)}{I_{q_\phi}(x; u | c)} +
        \ENB{q_\phi(c)} \E{q_\phi(u|c)}{\log \frac{q_\phi(u|c)}{p_\psi(u|c)}} \\
        &= \E{q_\phi(c)}{I_{q_\phi}(x; u | c)} +
        \E{q_\phi(c)}{\kl{q_\phi(u|c)}{p_\psi(u|c)}} \\
        &\ge \E{q_\phi(c)}{I_{q_\phi}(x; u | c)}
    \end{split}
\end{align*}

So $\ENB{p_D(x)}\E{q_\phi(c|x)}{\kl{q_\phi(u|c,x)}{p_\psi(u|c)}}$ is an upper bound on $\E{q_\phi(c)}{I_{q_\phi}(x; u | c)}$.

\subsection{Generalized Objective Function} \label{app:generalization}

We discussed that the model could be generalized to a multi-label setting,
\begin{gather*}
    \Vec{c} = \{c_1, c_2, \dots, c_K\} \\
    \phantom{.}\Vec{u} = \{u_1, u_2, \dots, u_K\}.
\end{gather*}

In this case, due to the independence assertions implied by the graphical model, equations
\begin{gather*}
    p(\Vec{c}) = \prod_{k=1}^K p(c_k) \\
    q_\phi(\Vec{c}|x) = \prod_{k=1}^K q_\phi(c_k|x) \\
    q_\phi(\Vec{u}|\Vec{c},x) = \prod_{k=1}^K q_\phi(u_k|\Vec{c},x) = \prod_{k=1}^K q_\phi(u_k|c_k,x) \\
    p_\psi(\Vec{u}|\Vec{c}) = \prod_{k=1}^K p_\psi(u_k|\Vec{c}) = \prod_{k=1}^K p_\psi(u_k|c_k)
\end{gather*}
hold, in which $K$ is the count of discrete variables.

We also redefine aggregate discrete posterior distribution, one for each discrete variable, as
\begin{equation*}
    \phantom{.}q_\phi(c_k) = \E{p_D(x)}{q_\phi(c_k|x)}.
\end{equation*}

With these extensions, we can rewrite the model's objective in Equation~\ref{eq.obj-capacity} as
\begin{align*}
    \mathcal{L}_\text{\modelname} &= 
    \ENB{p_D(x)} \ENB{q_\phi(\Vec{c}|x)} \ENB{q_\phi(\Vec{u}|\Vec{c},x)} \ENB{q_\phi(z|x)} \left[\log p_\theta(x|\Vec{c},\Vec{u},z)\right] \\
    &\quad - \gamma_u \; \E{p_D(x)}{\left| \sum_{k=1}^K
    \E{q_\phi(c_k|x)}{\kl{q_\phi(u_k|c_k,x)}{p_\psi(u_k|c_k)}}
    - C_u \right|} \\
    &\quad - \gamma_h \; \E{p_D(x)}{\sum_{k=1}^K H(q_\phi(c_k|x))} \\
    &\quad - \gamma_c \; \sum_{k=1}^K \kl{q_\phi(c_k)}{p(c_k)} \\
    &\quad - \gamma_z \; \E{p_D(x)}{|\kl{q_\phi(z|x)}{p(z)} - C_z|} \\
    &\quad - \gamma_{bc} \; \sum_{k=1}^K \sum_{i=1}^{L_k} \sum_{j=i}^{L_k} \max(BC(p_\psi(u_k|c_k=i), p_\psi(u_k|c_k=j)) - \delta, 0)
    ,
\end{align*}
in which, again, $K$ is the count of discrete variables, and $L_k$ is the number of possible values for $k$-th discrete variable.

\section{DATASETS}

We have experimented with three datasets:
\begin{itemize}
    \item MNIST (\cite{mnist}): 60,000, $28 \times 28$, grayscale images of handwritten digits;
    \item dSprites (\cite{dsprites}): 737,280, $64 \times 64$, binary images of 2D shapes;
    \item CelebA (\cite{celeba}): 202,599, $218 \times 178$, RGB images of celebrities faces. We used dataset's labels for \textit{Hair Color},
    \textit{Beard},
    \textit{Bang},
    \textit{Receding Hairline},
    \textit{Eyeglasses},
    \textit{Smiling}, and
    \textit{Gender}. We also labeled \textit{Skin Tone} on a fraction of training data points.
\end{itemize}

\section{MODEL ARCHITECTURE} \label{app:arch}

The architecture of model, for each dataset, is presented in Tables~\ref{app.table.mnist_arch}, \ref{app.table.dsprites_arch}, \ref{app.table.celeba_arch}. 

We use Sigmoid as the activation function of the decoder's last layer and after computing attention maps logits. A Softmax function is employed for calculating the discrete variables' distributions. In all other layers, we use ReLU (or Leaky ReLU) as the activation function.

Note that for MNIST, we resized input images to $32 \times 32$. In both MNIST and dSprites datasets, we used the same architecture as in \cite{jointvae}.

\begin{table}[h]
    \caption{Encoder and Decoder Architectures for the MNIST Dataset}
    \label{app.table.mnist_arch}
    \begin{center}
        \begin{tabular}{ll}
            \textbf{Encoder}    &\textbf{Decoder} \\
            \hline
            Input: 32 $\times$ 32 grayscale image & Input: Concat($z,u$) \\
            32 Conv. 4 $\times$ 4, stride 2 & FC. 256 \\
            64 Conv. 4 $\times$ 4, stride 2 & FC. 1024 \\
            64 Conv. 4 $\times$ 4, stride 2 & 32 Conv. Transpose 4 $\times$ 4, stride 2\\
            FC. 256 ($h$) & 32 Conv. Transpose 4 $\times$ 4, stride 2\\
            FC. 10 ($c$), FC. 2 $\times$ 6 ($z$) & 1 Conv. Transpose 4 $\times$ 4, stride 2 \\
            FC. 256 ($a$ from $c$) & \\
            FC. 2 $\times$ 10 ($u$ from $h.a$) & \\
        \end{tabular}
    \end{center}
\end{table}

\begin{table}[h]
    \caption{Encoder and Decoder Architectures for the dSprites Dataset}
    \label{app.table.dsprites_arch}
    \begin{center}
        \begin{tabular}{ll}
            \textbf{Encoder}    &\textbf{Decoder} \\
            \hline
            Input: 32 $\times$ 32 binary image & Input: Concat($z,u$) \\
            32 Conv. 4 $\times$ 4, stride 2 & FC. 256 \\
            32 Conv. 4 $\times$ 4, stride 2 & FC. 1024 \\
            64 Conv. 4 $\times$ 4, stride 2 & 64 Conv. Transpose 4 $\times$ 4, stride 2\\
            64 Conv. 4 $\times$ 4, stride 2 & 32 Conv. Transpose 4 $\times$ 4, stride 2\\
            FC. 256 ($h$) & 32 Conv. Transpose 4 $\times$ 4, stride 2\\
            FC. 3 ($c$), FC. 2 $\times$ 5 ($z$) & 1 Conv. Transpose 4 $\times$ 4, stride 2 \\
            FC. 256 ($a$ from $c$) & \\
            FC. 2 $\times$ 1 ($u$ from $h.a$) & \\
        \end{tabular}
    \end{center}
\end{table}

\begin{table}[h]
    \caption{Encoder and Decoder Architectures for the CelebA Dataset}
    \label{app.table.celeba_arch}
    \begin{center}
        \begin{tabular}{ll}
            \textbf{Encoder}    &\textbf{Decoder} \\
            \hline
            Input: 218 $\times$ 178 RGB image & Input: Concat($z,u$) \\
            32 Conv. 4 $\times$ 4, stride 2 & FC. 256 ($h$) \\
            32 Conv. 4 $\times$ 4, stride 2 & FC. 1600 \\
            64 Conv. 4 $\times$ 4, stride 2 & 64 Conv. Transpose 4 $\times$ 4, stride 2\\
            64 Conv. 4 $\times$ 4, stride 2 & 64 Conv. Transpose 4 $\times$ 4, stride 2\\
            64 Conv. 4 $\times$ 4, stride 2 & 32 Conv. Transpose 4 $\times$ 4, stride 2\\
            FC. 256 ($h$) & 32 Conv. Transpose 4 $\times$ 4, stride 2\\
            FC. 20 ($c$), FC. 2 $\times$ 10 ($z$) & 3 Conv. Transpose 4 $\times$ 4, stride 2 \\
            8 FC. 256 ($a$ from $c$ per disc. var.) & \\
            8 FC. 2 $\times$ 1 ($u$ from $h.a$ per disc. var.) & \\
        \end{tabular}
    \end{center}
\end{table}

\section{TRAINING DETAILS} \label{app:training}

We use PyTorch (\cite{pytorch}) to implement our model, and scikit-learn (\cite{sklearn}) to implement metrics. We train our model using Adam optimizer (\cite{adam}) with $\beta_1 = 0.9$, $\beta_2 = 0.999$, $\epsilon = 10^{-8}$, and different learning rates depending on the dataset. We also employ PyTorch's ReduceLROnPlateau, a learning rate scheduler, which reduces the learning rate if a metric stops improving. In all of our training models, we use a batch size of 64.
Additionally, we linearly increase $C_u$ and $C_z$ from $0$ to a specific number and in a particular number of iterations during the training. 
Table~\ref{app.table.training-specs} shows used parameters for different datasets.

\begin{table}[h]
    \caption{General Information and Hyperparameters of Models}
    \label{app.table.training-specs}
    \begin{center}
        \begin{tabular}{llll}
            \textbf{Characteristic} & \textbf{MNIST} & \textbf{dSprites} & \textbf{CelebA} \\
            \hline
            Number of discrete variables & 1 & 1 & 8 \\
            Number of classes of each disc. var. & [10] & [3] & [4, 3, 3, 2, 2, 2, 2, 2] \\ 
            Dimension of $u$ per disc. var. & 10 & 1 & 1 \\
            Dimension of $z$ & 6 & 5 & 10 \\
            Learning rate & 0.0015 & 0.0005 & 0.0005 \\
            $\gamma_c, \gamma_h$ & 15, 30 & 100, 10 & 2000, 10 \\
            $\gamma_z, \gamma_u$ & 15 & 50 & 1000 \\
            $\gamma_{bc}$ & 30 & 10 & 500 \\
            Capacity of $z$ ($C_z$) & 0 to 7 in 100000 iters & 0 to 30 in 300000 iters & 0 to 30 in 125000 iters\\
            Capacity of $u$ ($C_u$) & 0 to 7 in 100000 iters & 0 to 5 in 300000 iters & 0 to 15 in 125000 iters\\
            Intersection tolerance ($\delta$) & 0.15 & 0.1 & 0.2\\
            Reconstruction Error & BCE & BCE & MAE \\
            Epochs & 60 & 30 & 80 \\
        \end{tabular}
    \end{center}
\end{table}

\section{ADDITIONAL RESULTS}

In this section, we provide additional results on the dSprites dataset and further discuss the Bhattacharyya coefficient's role.

\subsection{Disentanglement on the dSprites Dataset}

\begin{figure}[h]
    \centering
    \begin{subfigure}[b]{0.30\textwidth}
        \centering
        \includegraphics[width=1\textwidth]{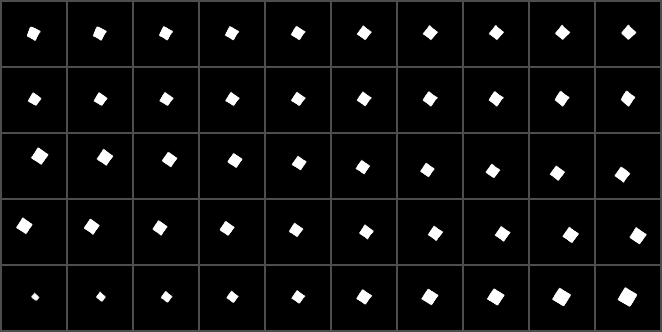}
        \caption{Square Class}
    \end{subfigure}
    \begin{subfigure}[b]{0.30\textwidth}
        \centering
        \includegraphics[width=1\textwidth]{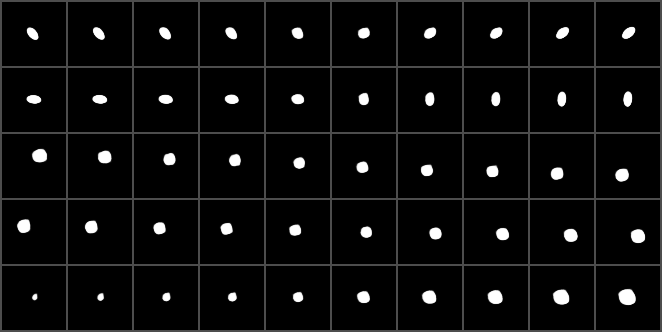}
        \caption{Ellipse Class}
    \end{subfigure}
    \begin{subfigure}[b]{0.30\textwidth}
        \centering
        \includegraphics[width=1\textwidth]{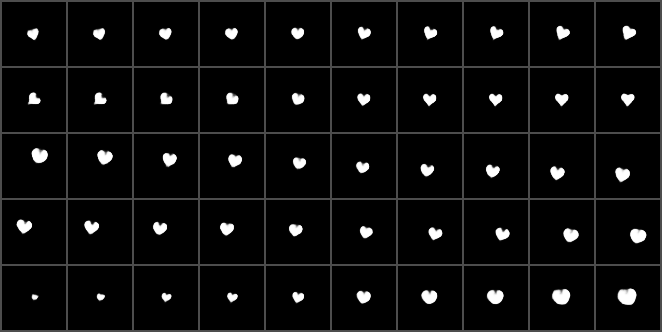}
        \caption{Heart Class}
    \end{subfigure}
    \caption{Latent Traversal of Class-Independent Factors on the dSprites dataset}
    \label{fig:dsprites_traversal}
\end{figure}

In Figure \ref{fig:dsprites_traversal}, we have presented qualitative results of a model trained on the dSprites dataset with $1.35\%$ supervision. This model has learned the 3 classes of shape and has disentangled them from the continuous variations, namely scale, orientation, and position.

\subsection{Learned Prior Distributions without Bhattacharyya Coefficient}  \label{app:eff-bc}

\begin{figure}[h]
    \centering
    \begin{subfigure}{\textwidth}
        \centering
        \includegraphics[width=1\textwidth]{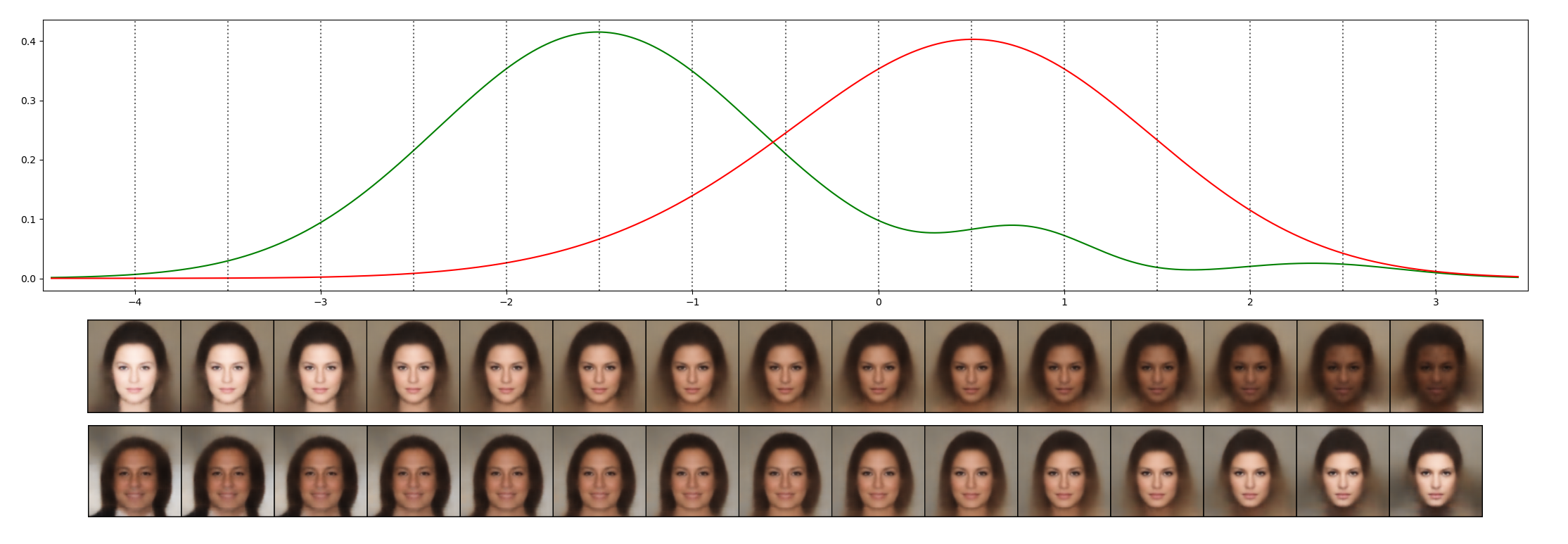}
        \caption{Distributions on \textit{Skin Tone} Factor}
    \end{subfigure}
    \begin{subfigure}{\textwidth}
        \centering
        \includegraphics[width=1\textwidth]{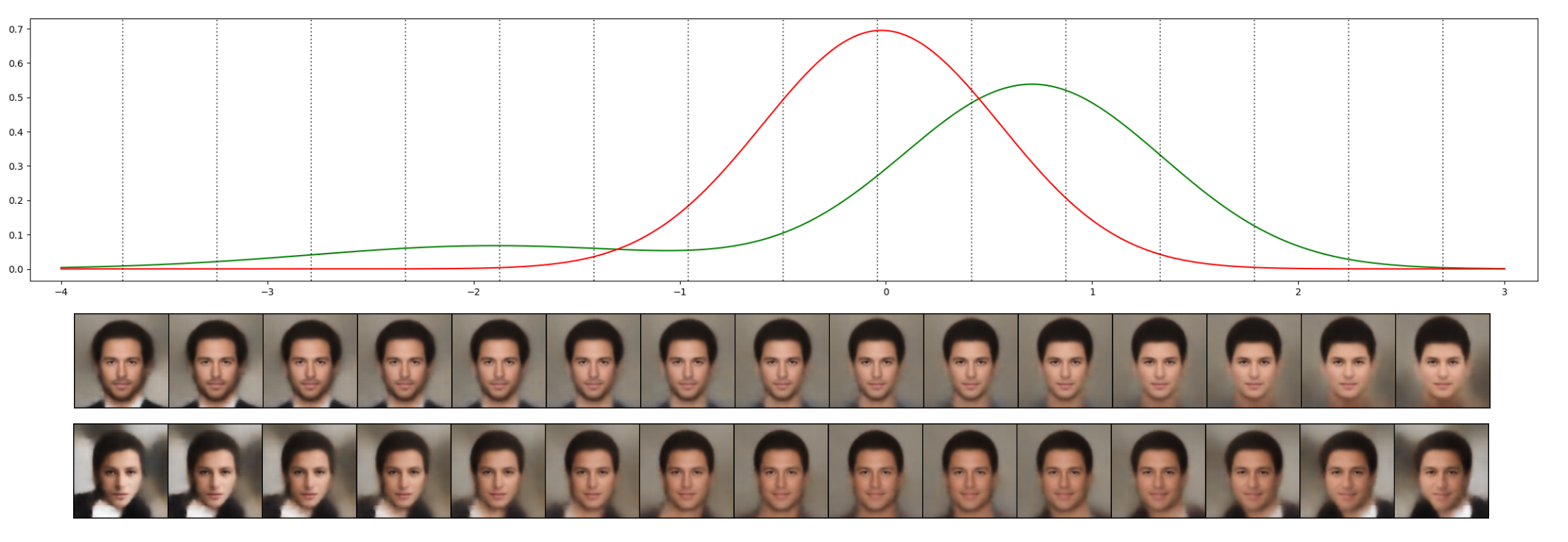}
        \caption{Distributions on \textit{Beard} Factor}
    \end{subfigure}
    \begin{subfigure}{\textwidth}
        \centering
        \includegraphics[width=1\textwidth]{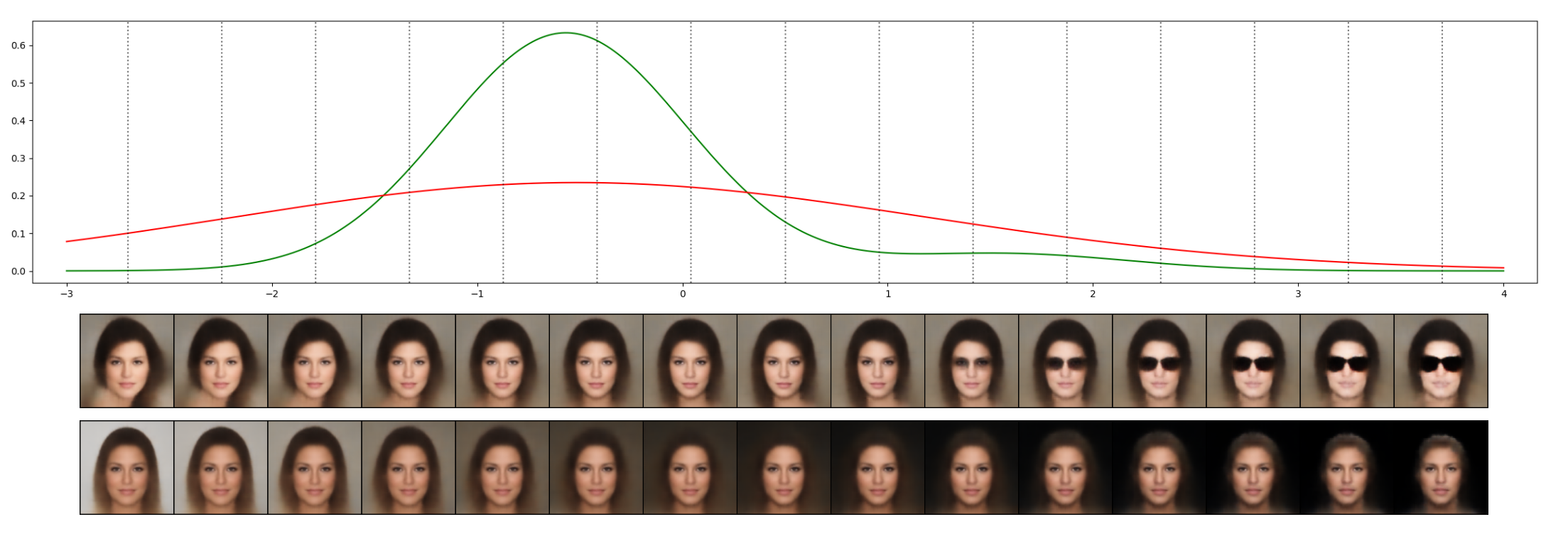}
        \caption{Distributions on \textit{Eyeglasses} Factor}
    \end{subfigure}
    \caption{Effect of Bhattacharyya Coefficient on the CelebA dataset. In every subfigure, on top, we have the mixture prior distributions for a specific factor. Green distributions are for the typical model, and red distributions are for the model without the Bhattacharyya coefficient. The second and third rows of every subfigure are generated samples from the typical model and the model without BC, respectively. In every face image, the corresponding element of $u$ is set to the value indicated by the black, dotted, vertical line in the distributions' plot. On the green distributions, multiple modes of the mixture are distinguishable (one mode is more apparent since the classes are imbalanced, and other modes are scaled by a small $p(c)$), but on the red distributions, there are no such distinctions between modes due to having similar means or large variances.}
    \label{app.fig:bc-comparison}
\end{figure}

To investigate whether the poor performance of models discussed in
Section~\ref{subsec:eff-bc} is due to inappropriate learning of prior distributions, we compare the mixture of Gaussians prior distributions of our typical model with a version of our model, lacking the Bhattacharyya coefficient. Figure~\ref{app.fig:bc-comparison} compares these distributions, as well as comparing samples generated from each of the models.
It implies that utilizing the Bhattacharyya coefficient has helped the generation performance of our model.

\end{document}